\documentclass{article}

\usepackage{microtype}
\usepackage{graphicx}
\usepackage{subfigure}
\usepackage{booktabs} 

\usepackage{hyperref}

\usepackage[accepted]{icml2020}

\icmltitlerunning{Storing Encoded Episodes as Concepts for Continual Learning}

\begin{document}

\twocolumn[
\icmltitle{Storing Encoded Episodes as Concepts for Continual Learning}



\begin{icmlauthorlist}
\icmlauthor{Ali Ayub}{1}
\icmlauthor{Alan R. Wagner}{2}
\end{icmlauthorlist}
\icmlaffiliation{1}{Department of Electrical Engineering, The Pennsylvania State University, State College, PA 16802, USA}
\icmlaffiliation{2}{Department of Aerospace Engineering, The Pennsylvania State University, State College, PA 16802, USA}

\icmlcorrespondingauthor{Ali Ayub}{aja5755@psu.edu}

\icmlkeywords{Continual Learning, Catastophic Forgetting, Cognitively-Inspired Learning}

\vskip 0.3in
]



\printAffiliationsAndNotice{}  

\begin{abstract}

 The two main challenges faced by continual learning approaches are catastrophic forgetting and memory limitations on the storage of data. To cope with these challenges, we propose a novel, cognitively-inspired approach which trains autoencoders with Neural Style Transfer to encode and store images. Reconstructed images from encoded episodes are replayed when training the classifier model on a new task to avoid catastrophic forgetting. The loss function for the reconstructed images is weighted to reduce its effect during classifier training to cope with image degradation. When the system runs out of memory the encoded episodes are converted into centroids and covariance matrices, which are used to generate pseudo-images during classifier training, keeping classifier performance stable with less memory. Our approach increases classification accuracy by \textbf{13-17\%} over state-of-the-art methods on benchmark datasets, while requiring \textbf{78\%} less storage space.
 
\end{abstract}
\section{Introduction}
\label{sec:introduction}
Humans continue to learn new concepts over their lifetime without the need to relearn most previous concepts. Modern machine learning systems, however, require the complete training data to be available at one time (batch learning) \cite{Girshick_2015_ICCV}. In this paper, we consider the problem of continual learning from the class-incremental perspective. Class-incremental systems are required to learn from a stream of data belonging to different classes and are evaluated in a single-headed evaluation \cite{Chaudhry_2018_ECCV}. In single-headed evaluation, the model is evaluated on all classes observed so far without any information indicating which class is being observed.

One of the main problems faced by class-incremental learning approaches is \textit{catastrophic forgetting} \cite{french19,kirkpatrick17}, in which the model forgets the previously learned classes when learning new classes and the overall classification accuracy decreases.
Most existing class-incremental learning methods avoid this problem by storing a portion of the training samples from previous classes and retraining the model (often a neural network) on a mixture of the stored data and new data~\cite{Rebuffi_2017_CVPR,Castro_2018_ECCV}. Storing real samples of the previous classes, however, leads to several issues. First, as pointed out by \cite{wu18}, storing real samples quickly exhausts memory capacity and limits performance for real-world applications. Second, storing real samples introduces privacy and security issues \cite{wu18}. Third, storing real samples is not biologically inspired, i.e. humans do not need to relearn previously known classes. 

In this paper we explore the "strict" class-incremental learning problem in which the model is not allowed to store any real samples of the previously learned classes. This problem has been previously addressed using generative models such as autoencoders or GANs (Generative Adversarial Networks) \cite{Seff17,Ostapenko_2019_CVPR,kemker18}. Most approaches for strict class-incremental learning use GANs to generate samples reflecting old class data \cite{Seff17,Wu18_NIPS,Ostapenko_2019_CVPR}, because GANs generate sharp, fine-grained images. The downside of GANs, however, is that they tend to generate images which do not belong to any of the learned classes, hurting classification performance. Autoencoders, on the other hand, always generate images that relate to the learned classes, but tend to produce blurry images that are also not good for classification.   

To cope with these issues, we propose a novel, cognitively-inspired approach termed Encoding Episodes as Concepts (EEC) for continual learning, which utilizes convolutional autoencoders to generate previously learned class data. Inspired by models of the hippocampus \cite{renoult15}, we use autoencoders to create compressed embeddings (encoded episodes) of real images and store them in memory. To avoid the generation of blurry images, we borrow ideas from the Neural Style Transfer (NST) algorithm \cite{Gatys_2016_CVPR} to train the autoencoders. For efficient memory management, we use the notion of \textit{memory integration}, from hippocampal and neocortical concept learning \cite{Mack18,moscovitch16}, to combine similar episodes into centroids and covariance matrices eliminating the need to store real data.

This paper contributes: 1) an autoencoder based approach to strict class-incremental learning which uses Neural Style Transfer to produce quality samples reflecting old class data (Sec. \ref{sec:autoencoder_training}); 2) a cognitively-inspired memory management technique that combines similar samples into a centroid/covariance representation, drastically reducing the memory required (Sec. \ref{sec:memory_integration}); 3) a data filtering and a loss weighting technique to manage image degradation of old classes during classifier training (Sec. \ref{sec:rehearsal}).
\section{Encoding Episodes as Concepts (EEC)}
\label{sec:methodlogy}
Following the notation of \cite{Chaudhry_2018_ECCV} and \cite{Ostapenko_2019_CVPR}, we consider $S_t = \{(x_i^t,y_i^t)\}_{i=1}^{n^t}$ to be the set of samples $x_i \in \mathcal{X}$ and their ground truth labels $y_i^t$ belonging to task $t$. In a class-incremental setup, data for different tasks is available to the model in smaller groups, hence $S_t$ can be composed of data belonging to one or multiple classes. In each increment the model is evaluated on all the labels (classes) seen so far. 

Formally, we follow the continual learning setup in \cite{Ostapenko_2019_CVPR}, where a task solver model (classifier for class-incremental learning) $D$ has to update its parameters $\theta_D$ on the data of task $t$ in an increment such that it performs equally well on all the $t-1$ previous tasks seen so far. Data for the $t-1$ tasks is not available when the model is learning task $t$. The main components of our approach to solve the class-incremental learning problem are mentioned below. 

\subsection{Autoencoder Training with Neural Style Transfer}
\label{sec:autoencoder_training}
An autoencoder is a neural network that is trained to compress and then reconstruct the input \cite{Goodfellow16}, formally $f_r: \mathcal{X}\rightarrow\mathcal{X}$. The network consists of an encoder responsible for compressing the input to a lower dimensional feature embedding (termed encoded episode in this paper) and a decoder that reconstructs the input from the feature embedding. 
The parameters $\theta_r$ of the network are usually optimized using an $l_2$ loss ($\mathcal{L}_r$) between the inputs and the reconstructions:

\begin{equation}
    \mathcal{L}_r = ||x-f_r(x)||_2
\end{equation}

Although autoencoders are suitable for dimensionality reduction for complex, high-dimensional data like RGB images, the reconstructed images lose the high frequency components necessary for correct classification. To tackle this problem, we utilize some of the ideas that underline Neural Style Transfer (NST) \cite{Gatys_2016_CVPR}, when training our autoencoders. NST uses a pre-trained CNN to transfer the style of one image to another. The process takes three images, an input image, a content image and a style image and alters the input image such that it has the content image's content and the artistic style of the style image. The three images are passed through the pre-trained CNN generating convolutional feature maps (usually from the last convolutional layer) and $l_2$ distances between the feature maps of the input image and content image (content loss) and style image (style loss) are calculated. These losses are then used to update the input image.

We only utilize the idea of content transfer from the full NST algorithm, where the input image is the image reconstructed by the autoencoder and content image is the real image corresponding to the reconstructed image. The classifier, D, is used as the pre-trained network to generate feature maps for the NST. This model has already been trained on real data for the classes in the increment \textit{t}. In contrast to the traditional NST algorithm, we use the content loss ($\mathcal{L}_{cont}$) to train the autoencoder, rather than updating the input image directly. Formally, let $f_c:\mathcal{X}\rightarrow\mathcal{F}_c$ be the classifier $D$ pipeline that converts input images into convolutional features. For an input image $x_i^t$ belonging to task $t$, the content loss is:

\begin{equation}
    \mathcal{L}_{cont} = ||f_c(x_i^t) - f_c(f_r(x_i^t))||_2
\end{equation}

The autoencoder parameters are optimized using a combination of reconstruction and content losses:

\begin{equation}
    \mathcal{L} = (1-\lambda)\mathcal{L}_r + \lambda\mathcal{L}_{cont}
\end{equation}

\begin{figure}[t]
\centering
\includegraphics[scale=0.33]{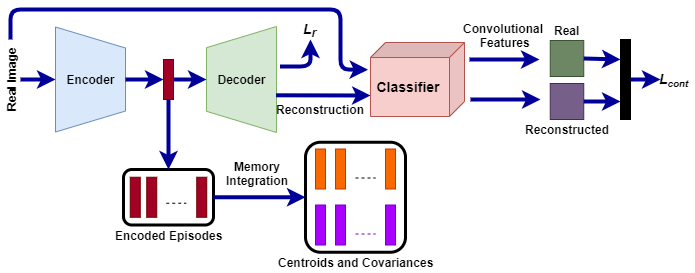}
\caption{\small For each new task, convolutional autoencoder takes real images as input and trained on a combination of reconstruction loss $\mathcal{L}_r$ and content loss $\mathcal{L}_{cont}$. The encoded episodes are stored in memory and converted into centroids and covariance matrices when the system runs out of memory.}
\label{fig:enc_dec}
\end{figure}

where, $\lambda$ is a hyperparamter that controls the contribution of each loss term towards the complete loss. During autoencoder training, classifier $D$ acts as a fixed feature extractor and its parameters are not updated. The complete procedure is depicted in Figure \ref{fig:enc_dec}. 

We train a separate autoencoder for each new task in an increment versus training a single autoencoder on the reconstructed images for all of the previous tasks. We utilize shallow convolutional autoencoders which require a fraction of the memory of the classifier $D$ ($\sim$0.4\%), hence the memory footprint does not grow drastically with each new increment. Still, if the system runs out of memory, a previous autoencoder can be re-utilized by training it on a combination of new task images and reconstructed images of the older tasks. In section \ref{sec:experiments} we present results using a single and multiple autoencoders.

\subsection{Memory Integration}
\label{sec:memory_integration}
For each new task $t$, the data is encoded and stored in memory. Even though the encoded episodes require less memory than the real images, the system can still run out of memory when managing a continuous stream of incoming tasks. To cope with this issue, we propose a process inspired by \textit{memory integration} in the hippocampus and the neocortex \cite{Mack18,renoult15,moscovitch16}. Memory integration combines a new episode with a previously learned episode summarizing the information in both episodes in a single representation.

Consider a system that can store a total of $K$ encoded episodes based on its available memory. Assume that at increment $t-1$, the system has a total of $K_{t-1}$ encoded episodes stored. It is now required to store $K_t$ more episodes in increment $t$. The system runs out of memory because $K_t + K_{t-1}>K$. Therefore, it must reduce the number of episodes to $K_r = K_{t-1} + K_t - K$. Because each task is composed of a set of classes at each increment, we reduce the total encoded episodes belonging to different classes based upon their previous number of encoded episodes. Formally, the reduction in the number of encoded episodes $N_y$ for a class $y$ is calculated as (whole number):

\begin{equation}
    N_{y}(new) = N_y(1-\frac{K_r}{K_{t-1}})
\end{equation}

To reduce the encoded episodes to $N_y(new)$ for class $y$, inspired by the memory integration process, we use an incremental clustering process similar to \textit{Agg-Var} clustering proposed in \cite{Ayub_scenes20,Ayub_2020_CVPR_Workshops}. The incremental clustering process combines the closest encoded episodes to produce centroids and covariance matrices. These centroids and covariance matrices (only diagonal entries) are of the same size as the encoded episodes and they notionally represent the summarized information of the clustered encoded episodes. The distance between episodes is calculated using Euclidean distance, and the centroids are calculated using the weighted mean of the episodes. The online clustering process is repeated until the sum of the total number of centroids, covariance matrices and the encoded episodes for class $y$ equal $N_y(new)$. The original episodes are removed and only the centroid and the covariance matrix are kept. Thus, our system keeps the memory used from growing while maintaining a stable level of performance.

\subsection{Rehearsal, Pseudorehearsal and Classifier Training}
\label{sec:rehearsal}
The classifier is trained on data from three sources: 1) the real data for task $t$, 2) reconstructed images generated from the autoencoder's decoder, and 3) pseudo-images generated from centroids and covariance matrices. Source (2) uses the encodings from the previous tasks to generate a set of reconstructed images by passing them through the autoencoder's decoder. This process is referred to as rehearsal \cite{Robins95}. 

If the system also has old class data stored as centroids/covariance matrix pairs, pseudorehearsal is employed \cite{Robins95}. For each centroid/covariance matrix pair of a task we sample a multivariate Gaussian distribution with mean as the centroid and the corariance matrix to generate a large set of pseudo-encoded episodes. The episodes are then passed through the autoencoder's decoder to generate pseudo-images for the previous classes. Many of the pseudo-images are noisy. To filter the noisy pseudo-images, we pass them through the model $D$, which has already been trained on the prior classes, to get predicted labels for each psuedo-image. We only keep those pseudo-images that have the same predicted label as the label of the centroid they originated from. We keep the same number of pseudo-images as the total number of encoded episodes represented by the centroid/covariance matrix pair.

The generated pseudo-images can still be quite different from the original image, hurting classifier performance. We therefore weigh the cross entropy loss term for the reconstructed stream and pseudo-images stream when training the classifier model $D$. For a previous task $t-1$, the weight $\Gamma_{t-1}$ (the sample decay weight) of the loss term $\mathcal{L}_{t-1}$ is defined as:

\begin{equation}
    \Gamma_{t-1} = e^{-\gamma\alpha_{t-1}}
\end{equation}

where, $\alpha_{t-1}$ represents the number of times an autoencoder has been trained on the pseudo-images or reconstructed images of task $t-1$ and $\gamma$ (the \textit{sample decay coefficient}) is a constant hyperparameter with value between 0 and 1 that controls the decay weight. The sample decay weight was calculated independently for each stream. Thus for a new increment, the total loss $\mathcal{L}_D$ for training $D$ on the reconstructed and pseudo-images of the old tasks and real images of the new task $t$ is defined as: 

\begin{equation}
    \mathcal{L}_D = \mathcal{L}_t + \sum_{i=1}^{t-1} (\Gamma_{i}^{r}\mathcal{L}_{i}^{r} + \Gamma_{i}^{p}\mathcal{L}_{i}^{p}) 
\end{equation}

where $\Gamma_{t-1}^{p}$ and ($\mathcal{L}_{t-1}^{p}$) are the sample decay weight and loss function for the pseudo-image stream and $\Gamma_{t-1}^{r}$ and ($\mathcal{L}_{t-1}^{r}$) are the sample decay weight and loss function for the reconstructed images. Thus, the loss value for the reconstructed and pseudo-images contribute less than the real images while training the classifier $D$, keeping performing competitively on all tasks.

\section{Experiments}
\label{sec:experiments}
We perform experiments on two benchmark datasets: MNIST \cite{Lechun98} and ImageNet-50 \cite{Russakovsky15} and compare our approach to state-of-the-art (SOTA) approaches.

\subsection{Datasets}
MNIST dataset comprises of grey-scale images of handwritten digits between 0 to 9, with 50,000 training images, 10,000 validation images and 10,000 test imgages. ImageNet-50 is a smaller subset of iLSVRC-2012 dataset containing 50 classes with 1300 training images and 50 validation images per class. All of the datasets' images are resized to 32$\times$32, similar to \cite{Ostapenko_2019_CVPR}.

\subsection{Implementaion Details}
We use the Pytorch deep learning framework \cite{torch19} for implementation and training of all neural network models. For MNIST dataset we use a 4-layer shallow convolutional autoencoder and for ImageNet-50 we use a 3-layer shallow convolutional autoencoder. Both autoencoders take about 0.2MB disk space for storage. For classification, for MNIST dataset we use same architecture as the discriminator of the 3-layer DCGAN \cite{Radford15} and for ImageNet-50 we use ResNet-18 \cite{He_2016_CVPR}.

Similar to DGMw \cite{Ostapenko_2019_CVPR}, we report top-1 average incremental accuracy on 5 and 10 classes ($A_5$ and $A_{10}$) for the MNIST dataset trained in an incremental fashion with one class per increment while for ImageNet-50 we report top-1 average incremental accuracy on 3 and 5 tasks/increments ($A_3$ and $A_{5}$) with 10 classes in each increment. For a fair comparison, we mainly compare against approaches with a generative memory replay component evaluated in a single-headed fashion. Among such approaches, to the best of our knowledge, DGMw \cite{Ostapenko_2019_CVPR} represents the state-of-the-art benchmark on these datasets which is followed by MeRGAN \cite{Wu18_NIPS}, DGR \cite{Shin17} and EWC-M \cite{Seff17}. Joint training (JT) is used as an upperbound for both of the datasets. We compare all of these methods with two variants of our approach: EEC in which we use separate autoencoders for images available in different increments and EECs in which we use a single autoencoder and keep retraining it on the reconstructed images of the old classes images when new class data is available. For both of our models (EEC and EECs) we report results when all the encoded episodes of the previous classes are stored. 

Hyperparameters $\lambda$, $\gamma^r$ (for reconstructed images) and $\gamma^p$ (for psuedo-images) were chosen using cross-validation. We performed each experiment a total of 10 times with different random seeds and report average accuracy over all the runs. 

\begin{table}[t]
\centering
\small
\begin{tabular}{p{3.3cm}p{0.7cm}p{0.7cm}p{0.7cm}p{0.7cm}}
     \hline
     \multicolumn{1}{c}{} & \multicolumn{2}{c}{\textbf{MNIST}} & \multicolumn{2}{c}{\textbf{ImageNet-50}}\\
     \hline
    \textbf{Methods} & $A_5$(\%) & $A_{10}$(\%) & $A_3$(\%) & $A_{5}$(\%) \\
     \hline
    JT & 99.87 & 99.24 & 57.35 & 49.88 \\
     \hline
     iCaRL-S \cite{Rebuffi_2017_CVPR} & 84.61 & 55.8 & 29.38 & 28.98 \\
     EWC-S \cite{kirkpatrick17} & - & 79.7 & - & -\\
     RWalk-S \cite{Chaudhry_2018_ECCV} & - & 82.5 & - & - \\
     EEIL-S \cite{Castro_2018_ECCV} & - & - & 27.87 & 11.80\\
     \hline
     EWC-M \cite{Seff17} & 70.62 & 77.03 & - & -\\
     DGR \cite{Shin17} & 90.39 & 85.40 & - & -\\
     MeRGAN \cite{Wu18_NIPS} & 99.15 & 96.83 & - & -\\
     DGMw \cite{Ostapenko_2019_CVPR} & 98.75 & 96.46 & 32.14 & 17.82\\
     \hline
     EEC (Ours) & \textbf{99.20} & \textbf{97.83} & \textbf{45.39} & \textbf{35.24}\\
     Difference & \textbf{+0.05} & \textbf{+1.00} & \textbf{+13.2} & \textbf{+17.4}\\
     EECS (Ours) & 98.00 & 96.26 & 41.13 & 30.89\\
 \hline
 \end{tabular}
 \caption{Comparison of EEC and EECs against two different types of benchmarks: those that use episodic memory (real examples) and those that use generative memory (S denotes episodic memory approaches), for a class-incremental learning task on MNIST and ImageNet-50 datasets. The difference between EEC and the best SOTA approach is also shown.}
 \label{tab:comparison_with_sota}
 \end{table}

\subsection{Comparison with SOTA methods}
Table \ref{tab:comparison_with_sota} compares the two variants of our method against state-of-the-art (SOTA) approaches on the MNIST and ImageNet-50 datasets. We compare against two different types of approaches, those that use real images ("episodic memory") of the old class data when learning new classes and others that use generative memory to generate previous class data when learning new classes. Our methods (EEC and EECS) outperform the SOTA methods EWC-M and DGR by significant margins on the MNIST dataset for the 5 and 10 task ($A_5$ and $A_{10}$) experiments. MeRGAN and DGMw perform similarly to our methods on the $A_5$ and $A_{10}$ experiments. MeRGAN and EEC approach the JT upperbound on $A_5$ suggesting that this number of increments results in minimal catastrophic forgetting. Further, the accuracy for MeRGAN, DGMw and our method changes only slightly between $A_5$ and $A_{10}$, suggesting that MNIST is, perhaps, too simple of a dataset for testing continual learning using generative replay.


We now consider the more complex ImageNet-50 dataset. Only DGMw \cite{Ostapenko_2019_CVPR} reported results for this dataset. On ImageNet-50, both EEC and EECS outperform DGMw on $A_3$ and $A_5$ by significant margins (\textbf{13.29\%} and \textbf{17.42\%}, respectively). The accuracy achieved by EEC on $A_5$ is even higher than DGMw's accuracy on $A_3$. Further, EEC also beats iCaRL (episodic memory SOTA method) with margins of \textbf{16.01\%} and \textbf{6.26\%} on $A_3$ and $A_5$, respectively, even though iCaRL has an unfair advantage of using stored real images.

\begin{figure}[t]
\centering
\includegraphics[scale=0.6]{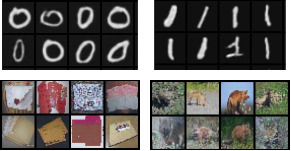}
\caption{\small Images reconstructed by EEC for MNIST (top) and ImageNet-50 (bottom) after all tasks.}
\label{fig:reconstructions}
\end{figure}

It is worth noting that our method performed reasonably consistently across both datasets. In contrast, DGMw (the best current method) shows significantly different results across both datasets. The results suggest that the current generative memory based SOTA approaches are unable to mitigate catastrophic forgetting on more complex RGB datasets. This could be because GANs tend to generate images that do not belong to any of learned classes, which can drastically reduce classifier performance. Our approach copes with these issues by training autoencoders with the ideas from the NST algorithm and retraining of the classifier with sample decay weights. Images reconstructed by EEC for MNIST after 10 tasks and ImageNet-50 after 5 tasks are shown in Figure \ref{fig:reconstructions}.

\paragraph{Memory Usage Analysis.} 
Similar to \cite{Ostapenko_2019_CVPR}, we analyze the disc space required by our model for the ImageNet-50 dataset. For EEC, the autoencoders use a total disc space of 1 MB, ResNet-18 uses about 44 MB, while the encoded episodes use a total of about 
66.56 MB. Hence, the total disc space required by EEC is about 111.56 MB. DGMw's generator (with corresponding weight masks), however, uses 228MB of disc space and storing pre-processed real images of ImageNet-50 requires disc space of 315MB. Hence, our model requires \textbf{51.07\%} ((228-111.56)/228 = 0.5107) less space than DGMw and \textbf{64.58\%} less space than the real images for ImageNet-50 dataset. 

To evaluate the effect of cognitively inspired memory management technique, we tested EEC with a memory budget of K=5000, where K is the sum of the total number of encoded episodes, centroids and covariance matrices (diagonal entries) stored by the system. For K=5000, on $A_3$ and $A_5$, EEC achieves \textbf{42.29\%} and \textbf{31.51\%} accuracies, respectively, which are only \textbf{3.1\%} and \textbf{3.73\%} lower than the accuracy of EEC with unlimited memory. Further, even for K=5000, EEC beats DGMw (current SOTA on ImageNet-50) by margins of \textbf{10.17\%} and \textbf{13.73\%} on $A_3$ and $A_5$, respectively. The total disc space required for K=5000 is only 5.12 MB and the total disc space for the complete system is 50.12 MB (44 MB for ResNet-18 and 1 MB for autoencoders), which is \textbf{78.01\%} less than DGMw's required disc space (228 MB). These results clearly depict that our approach produces best results even with extremely limited memory, a trait that is not shared by other SOTA approaches. Moreover, the results also show that our approach is capable of dealing with the two main challenges of continual learning mentioned earlier: catastrophic forgetting and memory management.

\section{Conclusion}
\label{sec:conclusion}

This paper has explored a new and potentially powerful approach to strict class-incremental learning we call Encoding Episodes as Concepts (EEC). Our paper demonstrates that the generation of high quality reconstructed data can serve as the basis for improved classification during continual learning. We further demonstrate techniques for dealing with image degradation during classifier training on new tasks. We have also demonstrated that a clustering approach can be used to manage the memory used by our system. Together, our experimental results demonstrate that these techniques mitigate the effects of catastrophic forgetting, especially on complex RGB datasets, while also using less memory that other SOTA approaches. As such, this work may offer a variety of potential avenues for future research, such as further improving data recreation accuracy or applying this approach to diverse, interesting problems. Future continual learning approaches can incorporate different components of our approach such as the NST based autoencoder, psuedo-rehearsal and sample decay weights for improved performance. 

\section*{Acknowledgments}
\noindent This work was supported by Air Force Office of Scientific Research contract FA9550-17-1-0017.


\bibliography{main}
\bibliographystyle{icml2020}

\end{document}